\documentclass[pdflatex,sn-nature]{sn-jnl}

\usepackage{graphicx}%
\usepackage{multirow}%
\usepackage{amsmath,amssymb,amsfonts}%
\usepackage{amsthm}%
\usepackage{mathrsfs}%
\usepackage[title]{appendix}%
\usepackage{xcolor}%
\usepackage{textcomp}%
\usepackage{manyfoot}%
\usepackage{booktabs}%
\usepackage{algorithm}%
\usepackage{algorithmicx}%
\usepackage{algpseudocode}%
\usepackage{listings}%
\usepackage{stfloats}
\usepackage{CJKutf8}
\usepackage{tabularx}
\usepackage{makecell}
\usepackage{array}
\usepackage{booktabs}
\usepackage{caption}
\usepackage{longtable}

\usepackage{./style/trackchanges}

\newcolumntype{L}{>{\raggedright\arraybackslash}X}

\theoremstyle{thmstyleone}%
\theoremstyle{thmstyletwo}%

\theoremstyle{thmstylethree}%

\begin{document}
\begin{CJK}{UTF8}{gbsn}

\title{BioMARS: A Multi-Agent Robotic System for Autonomous Biological Experiments}


\author[1,2]{\fnm{Yibo} \sur{Qiu}}\email{alexandreqiu@mail.ustc.edu.cn}
\equalcont{These authors contributed equally to this work.}

\author[1,2]{\fnm{Zan} \sur{Huang}}\email{huangzan@mail.ustc.edu.cn}
\equalcont{These authors contributed equally to this work.}

\author[1,2]{\fnm{Zhiyu} \sur{Wang}}\email{wzy2002@email.ustc.edu.cn}

\author[1,2]{\fnm{Handi} \sur{Liu}}\email{lhd24916055@mail.ustc.edu.cn}

\author[1,2]{\fnm{Yiling} \sur{Qiao}}\email{qiaoyilin1997@mail.ustc.edu.cn}

\author[1,2]{\fnm{Yifeng} \sur{Hu}}\email{yifengmed@mail.ustc.edu.cn}

\author[1,2]{\fnm{Shu'ang} \sur{Sun}}\email{shuangsun040901@gmail.com}

\author[1,2]{\fnm{Hangke} \sur{Peng}}\email{penghangke@gmail.com}

\author*[1,2]{\fnm{Ronald X} \sur{Xu}}\email{xux@ustc.edu.cn}

\author*[1,2]{\fnm{Mingzhai} \sur{Sun}}\email{mingzhai@ustc.edu.cn}

\affil*[1,2]{\orgdiv{Suzhou Institute for Advanced Research}, \orgname{University of Science and Technology of China}, \orgaddress{\city{Suzhou}, \state{Jiangsu}, \country{China}}}
\affil*[2]{\orgdiv{School of Biomedical Engineering},\orgdiv{Division of Life Sciences and Medicine},\orgname{University of Science and Technology of China},\orgaddress{\city{Hefei},\state{Anhui},\country{China}}}






\abstract{
Large language models (LLMs) and vision–language models (VLMs) have the potential to transform biological research by enabling autonomous experimentation \cite{o2023bioplanner, huang2024crispr}. Yet, their application remains constrained by rigid protocol design, limited adaptability to dynamic lab conditions, inadequate error handling \cite{cooper2025lira}, and high operational complexity.  Here we introduce BioMARS (Biological Multi-Agent Robotic System), an intelligent platform that integrates LLMs, VLMs, and modular robotics to autonomously design, plan, and execute biological experiments.  BioMARS uses a hierarchical architecture: the Biologist Agent synthesizes protocols via retrieval-augmented generation;  the Technician Agent translates them into executable robotic pseudo-code;  and the Inspector Agent ensures procedural integrity through multimodal perception and anomaly detection.  The system autonomously conducts cell passaging and culture tasks, matching or exceeding manual performance in viability, consistency, and morphological integrity. It also supports context-aware optimization, outperforming conventional strategies in differentiating retinal pigment epithelial cells.  A web interface enables real-time human–AI collaboration, while a modular backend allows scalable integration with laboratory hardware.  These results highlight the feasibility of generalizable, AI-driven laboratory automation and the transformative role of language-based reasoning in biological research.
}




\maketitle
\section{Introduction}\label{sec:Intorduction}
The convergence of robotic automation and artificial intelligence is reshaping experimental biology, promising greater reproducibility, throughput, and independence from human variability \cite{holland2020automation}. However, the complexity of biological protocols—which demand adaptive decision-making, multi-stage coordination, and interpretation of nuanced environmental feedback—has hindered the realization of fully autonomous systems. Existing automation solutions, ranging from specialized liquid handling robots \cite{dettinger2022open, novak2020robotic, taguchi2023automation}, to modular single-arm platforms for cell culture automation \cite{hamm2024modular, tristan2021robotic}, and dual-arm platforms enabling automated cell production \cite{koniger2024rebia, yachie2017robotic, ochiai2021variable}, often require extensive manual oversight and lack the flexibility to navigate unanticipated procedural deviations. Early systems focused on streamlining specific tasks, including biofoundries \cite{chao2017engineering}, IoT-enabled experimental platforms \cite{miles2018achieving}, and clinical sample preparation \cite{muller2020automated}, but these non-robotic arm systems still faced hardware limitations that prompted the development of robotic arm solutions.

Concurrently, large language models (LLMs) and vision–language models (VLMs) are transforming scientific problem-solving by enabling machines to parse literature, synthesize knowledge, and execute multi-modal reasoning across diverse domains \cite{vaswani2017attention, wang2024biorag, luu2024bioinspiredllm, zhang2025scientific}. Recent efforts leveraging LLMs in chemical experimentation \cite{boiko2023autonomous, darvish2025organa, cooper2025lira} and biological protocol generation \cite{o2023bioplanner, huang2024crispr} signal a paradigm shift toward AI-native experimentation. Yet, their integration with physical robotic systems for biological execution remains underexplored.

Here we introduce BioMARS (Biological Multi-Agent Robotic System), a dual-arm robotic platform orchestrated by LLMs and VLMs \cite{zhu2023minigpt, zhang2024vision} for fully autonomous execution of biological experiments. BioMARS performs end-to-end protocol design, environmental coordination, and robotic manipulation through adaptive multimodal reasoning. By converting research literature into actionable procedures and coupling them with error-aware execution strategies, the system ensures both flexibility and robustness in complex biological tasks.

We demonstrate BioMARS across five experimental capabilities: (1) efficiently searching and analyzing online research documentation to design experimental protocols for diverse cell types under varying conditions; (2) accurately translating and executing these protocols using a dual-arm biological laboratory; (3) detecting experimental errors via keyframe analysis; (4) performing end-to-end cell culturing; and (5) resolving optimization issues through the analysis of historical experimental data. 
\section{Result}
\label{sec:Result}

\subsection{Architecture of BioMARS system}
\label{sec:Architecture of BioMARS system}

\begin{figure*}[ht]
\centering
\includegraphics[width=0.9\textwidth]{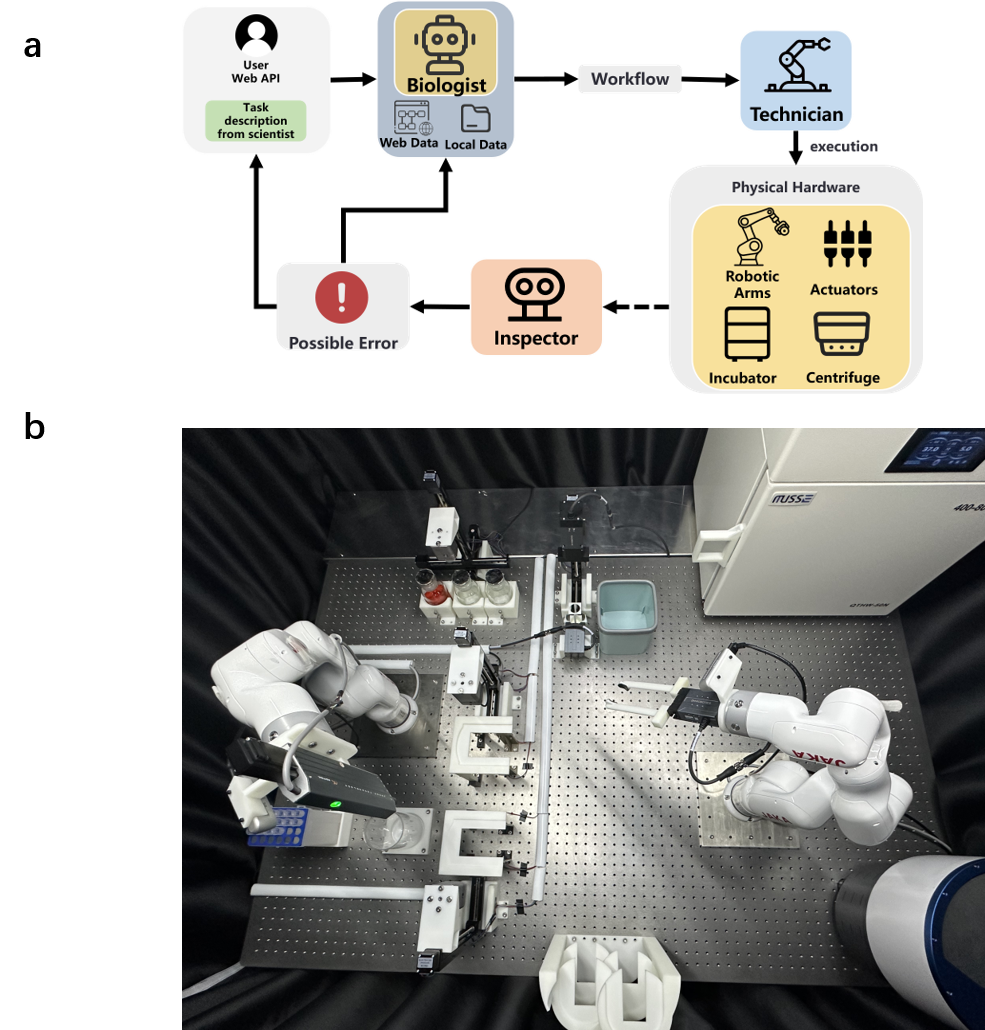}
\caption{\textbf{System architecture and robotic setup.}  
\textbf{a,} Multi-agent workflow of BioMARS, comprising Biologist, Technician, and Inspector agents. 
\textbf{b,} Dual-arm robotic platform configured for autonomous biological experimentation. }
\label{fig:structure}
\end{figure*}

BioMARS (Biological Multi-Agent Robotic System) enables end-to-end autonomous execution of biological experiments through a network of specialized LLM- and VLM-based agents (Fig.~\ref{fig:structure}a). Built on an enhanced Agentic Retrieval-Augmented Generation (RAG) framework with modular error correction \cite{singh2025agentic}, BioMARS decomposes complex protocols, interprets unstructured literature, and dynamically synthesizes findings into executable procedures.

The Biologist Agent ingests diverse open-access research documents, generates executable protocol steps by leveraging biological domain knowledge to create structured, constraint-aware queries. By incorporating constraints such as container type (e.g., petri dishes, flasks) and platform capacity, it tailors each protocol to the laboratory’s operational environment. The Technician Agent transforms high-level plans into fine-grained control primitives for robotic execution. These primitives are allocated across dual robotic arms and coordinated with environmental modules such as incubator and centrifuge.

To ensure execution robustness, the Inspector Agent—powered by ViTs and VLMs—performs rapid anomaly detection. It identifies procedural deviations including geometric misalignments (e.g., unattached pipette tips, misaligned petri dishes) and mechanical failures, prompting replanning or user notification. This tri-agent system mirrors the modularity and task specialization seen in other autonomous platforms \cite{boiko2023autonomous, m2024augmenting}, enabling BioMARS to operate adaptively under changing experimental conditions.

The platform supports natural language prompts (e.g., “How to passage HeLa cells”) via a web interface. Users can initiate, monitor, and modify experiments interactively. Critically, BioMARS’s modular architecture allows seamless integration of new hardware and protocol domains through programmable function modules, facilitating extensibility across diverse biological workflows.

\subsection{Protocol Synthesis under Environmental Constraints}
\label{Protocol Synthesis under Environmental Constraints}
\begin{figure*}[htbp]
\centering

\centering
\includegraphics[width=0.9\textwidth]{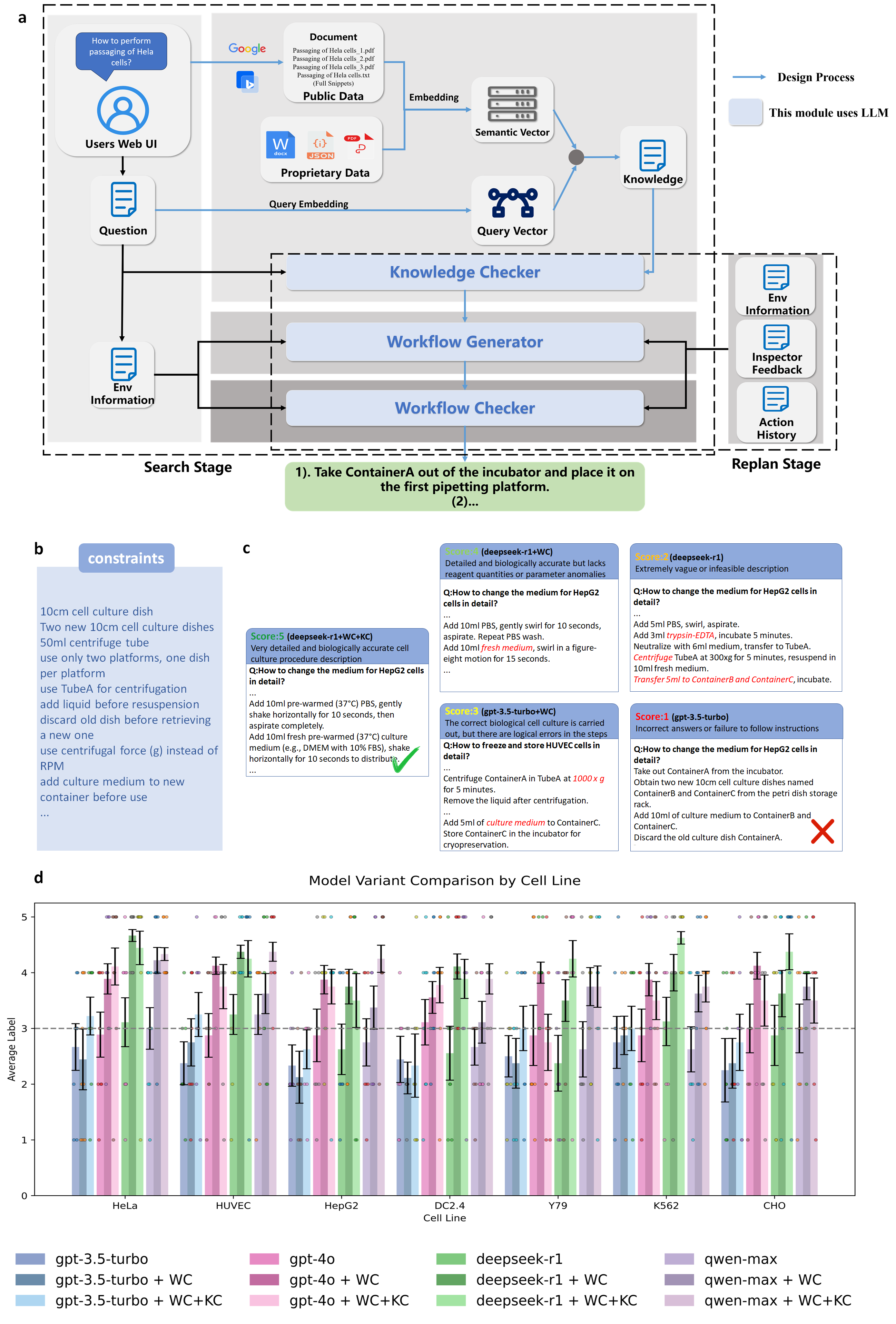}
\caption{\textbf{Biologist Agent architecture and evaluation.} \textbf{a,} Biologist agent pipeline integrating document retrieval, semantic matching, and workflow refinement under constraints.
\textbf{b,}  Representative experimental constraints. 
\textbf{c,}  Example protocol outputs with scores and errors.
\textbf{d,}  Performance comparison of four models (GPT-3.5-Turbo, GPT-4o, Deepseek-R1, Qwen-Max) and their variants on seven cell lines.
}
\label{fig:biologist_agent}
\end{figure*}

Reliable generation of biological protocols from literature poses challenges due to procedural complexity, heterogeneous experimental conditions, instrumentation constraints, and output formatting requirements. BioMARS addresses this through a multi-agent reasoning framework that integrates LLM-based planning with vector-based retrieval and verification mechanisms to generate biologically accurate, context-aware procedures.

At the center of this system is the Biologist Agent, which operates within an enhanced Agentic Retrieval-Augmented Generation (RAG) architecture (Fig.~\ref{fig:biologist_agent}a). The agent retrieves relevant knowledge using online query APIs (Google and Bing), extracting three PDFs and three high-relevance web snippets per query. Full paragraphs associated with each snippet are selected to preserve semantic context. These passages, along with the embedded user query (using OpenAI’s Ada model), undergo vector similarity ranking. The top five text chunks are used as context for downstream protocol generation.

Protocol construction is distributed across three sub-agents: the Knowledge Checker (KC), which filters domain-inconsistent content; the Workflow Generator (WG), which formulates stepwise procedures; and the Workflow Checker (WC), which iteratively refines outputs for logical coherence. The system accounts for laboratory constraints, such as limited stock of specific containers (e.g., 10 cm culture dishes), pipette tip volume (≤10 mL), and robotic station limits, ensuring all outputs are executable on the BioMARS platform.

System performance was evaluated using a 70-query benchmark (Appendix~\ref{secB1:10-Query Biological Task Set}) comprising 10 procedural categories across seven cell lines, ranging from routine tasks (e.g., cell passaging, thawing) to complex protocols (e.g., 3D culture, apoptosis analysis). Following Boiko et al. \cite{boiko2023autonomous}, model outputs were scored on a 5-point scale: 5 for fully detailed and accurate procedures; 4 for biologically sound steps with minor omissions; 3 for logically flawed but conceptually plausible outputs; 2 for vague or infeasible workflows; and 1 for incorrect or non-compliant procedures. Outputs below a score of 3 were considered task failures. Fig.~\ref{fig:biologist_agent}c presents representative outputs with annotations.

Without WC or KC modules, base models—including GPT-4o, Qwen-Max, and DeepSeek-R1—did not exceed a mean score of 3. GPT-3.5 Turbo consistently underperformed; in one instance, it misinterpreted “How to change the HepG2 culture medium” by suggesting disposal of viable dishes and initiating culture from scratch (score: 1). DeepSeek-R1 proposed cell redistribution via trypsinization (score: 2), demonstrating procedural confusion.

Incorporation of the WC module significantly improved structural logic. For example, DeepSeek-R1+WC successfully outlined PBS rinsing and medium replacement steps but omitted critical conditions (temperature, CO₂ levels), yielding a score of 4. Further integration with the KC module provided domain-specific validations: in the cryopreservation task for HUVECs, KC-corrected protocols mitigated centrifugation errors and ensured cryostorage in liquid nitrogen.

The best-performing configuration—DeepSeek-R1+WC+KC—achieved consistent scores of 5. Its output for HepG2 medium replacement detailed exact reagent volumes, environmental settings (37$^\circ$C, 5\% CO₂), and handling protocols (PBS rinse with horizontal agitation), aligning closely with expert protocols. These results affirm the critical role of domain validation (KC) and procedural refinement (WC) in transforming LLM outputs into executable, high-fidelity biological protocols (Fig.~\ref{fig:biologist_agent}c,d).

\subsection{Protocol-to-Code Translation for Robotics}
\label{Protocol-to-Code Translation for Robotics}




\begin{figure*}[htbp]
\centering
\includegraphics[width=0.9\textwidth]{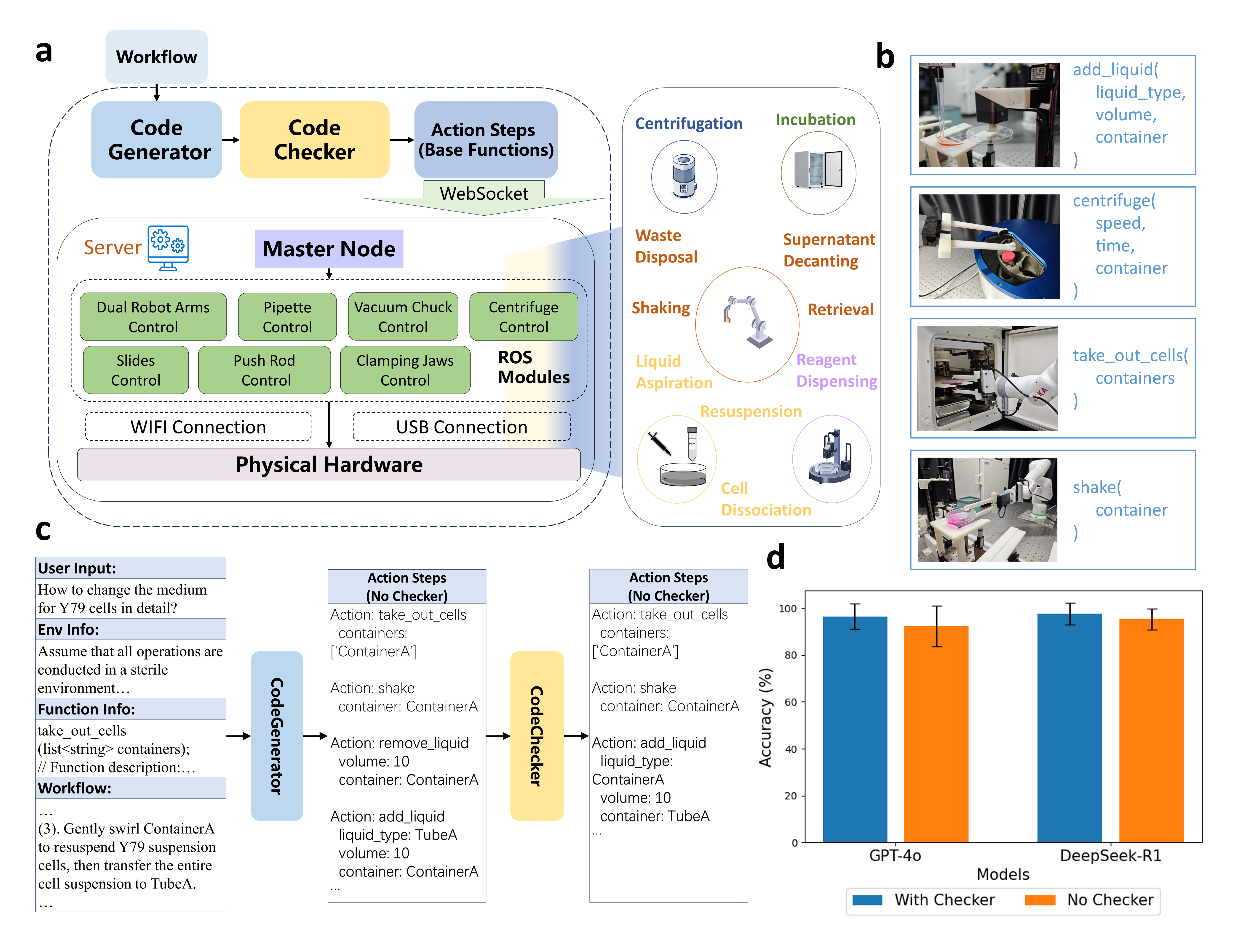}

\caption{\textbf{Technician Agent architecture and performance of protocol translation and execution framework.}  \textbf{a,} System workflow of Technician Agent, including CodeGenerator, CodeChecker, ROS node and the corresponding hardware module.
\textbf{b,} Example pseudo-code instructions and corresponding robotic actions.
\textbf{c,} The specific workflow of Technician Agent.
\textbf{d,} Instruction accuracy comparison with and without CodeChecker for GPT-4o and DeepSeek-R1.}
\label{fig:assistant_agent}
\end{figure*}

Translating free-text experimental protocols into executable robotic commands remains a central bottleneck in laboratory automation. Existing systems typically rely on rigid, manually curated command sequences \cite{kanda2022robotic,koniger2024rebia}, which limits their adaptability to diverse and unstructured inputs. To address this constraint, we developed the Technician Agent—a dual-module system that autonomously interprets natural language protocols and converts them into validated robotic instructions.

The Technician Agent operates through a cooperative pipeline comprising a CodeGenerator and a CodeChecker module (Fig.~\ref{fig:assistant_agent}a). The CodeGenerator, powered by an LLM, maps protocol descriptions into pseudo-code composed of primitive robotic operations such as \texttt{add\_liquid}, \texttt{centrifuge}, and \texttt{shake} (Fig.~\ref{fig:assistant_agent}b). The CodeChecker subsequently performs rule-based validation, enforcing functional correctness and environmental compatibility based on the predefined specification set (Appendix~\ref{secC1:Functional Interface Specifications}).

The pipeline structure is illustrated in Fig.~\ref{fig:assistant_agent}c. Given a protocol input, the CodeGenerator produces candidate instructions tailored to the lab environment. These instructions are then parsed by the CodeChecker, which applies logical and semantic checks including parameter validation, function relevance, and argument structure. This ensures that all generated commands adhere to the operational and safety constraints of the BioMARS platform.

To assess performance, we benchmarked the Technician Agent across 300 experimental protocol steps. As shown in Fig.~\ref{fig:assistant_agent}d, the full pipeline (CodeGenerator + CodeChecker, GPT-4o) achieved a 96.4\% instruction-matching accuracy, outperforming a single-module baseline (92.4\%). The impact of the CodeChecker module is particularly evident in complex procedural constructs. For example, when parsing the instruction “resuspend the cell pellet in 10 mL fresh complete growth medium,” the baseline failed to recognize the prerequisite transfer step. In contrast, the Technician Agent inserted an implicit \texttt{add\_liquid} operation before resuspension, preserving procedural logic.

Beyond resolving implicit steps, the CodeChecker module also corrects parameter mismatches, enforces range constraints, and eliminates superfluous instructions. For instance, it detects and corrects overfilled volumes relative to container capacity and replaces invalid data types in function arguments. This systematic refinement substantially improves the robustness of the robotic instruction set.

By converting ambiguous natural language into explicit, verifiable pseudo-code, the Technician Agent enhances experimental reproducibility, reduces human error, and simplifies execution on robotic platforms. This capability shifts the experimental burden away from manual coding, enabling researchers to focus on scientific inquiry rather than operational encoding.

\subsection{Hierarchical VLM-Based Error Detection}
\label{Hierarchical VLM-Based Error Detection}
\begin{figure*}[htbp]
\centering
\includegraphics[width=0.9\textwidth]{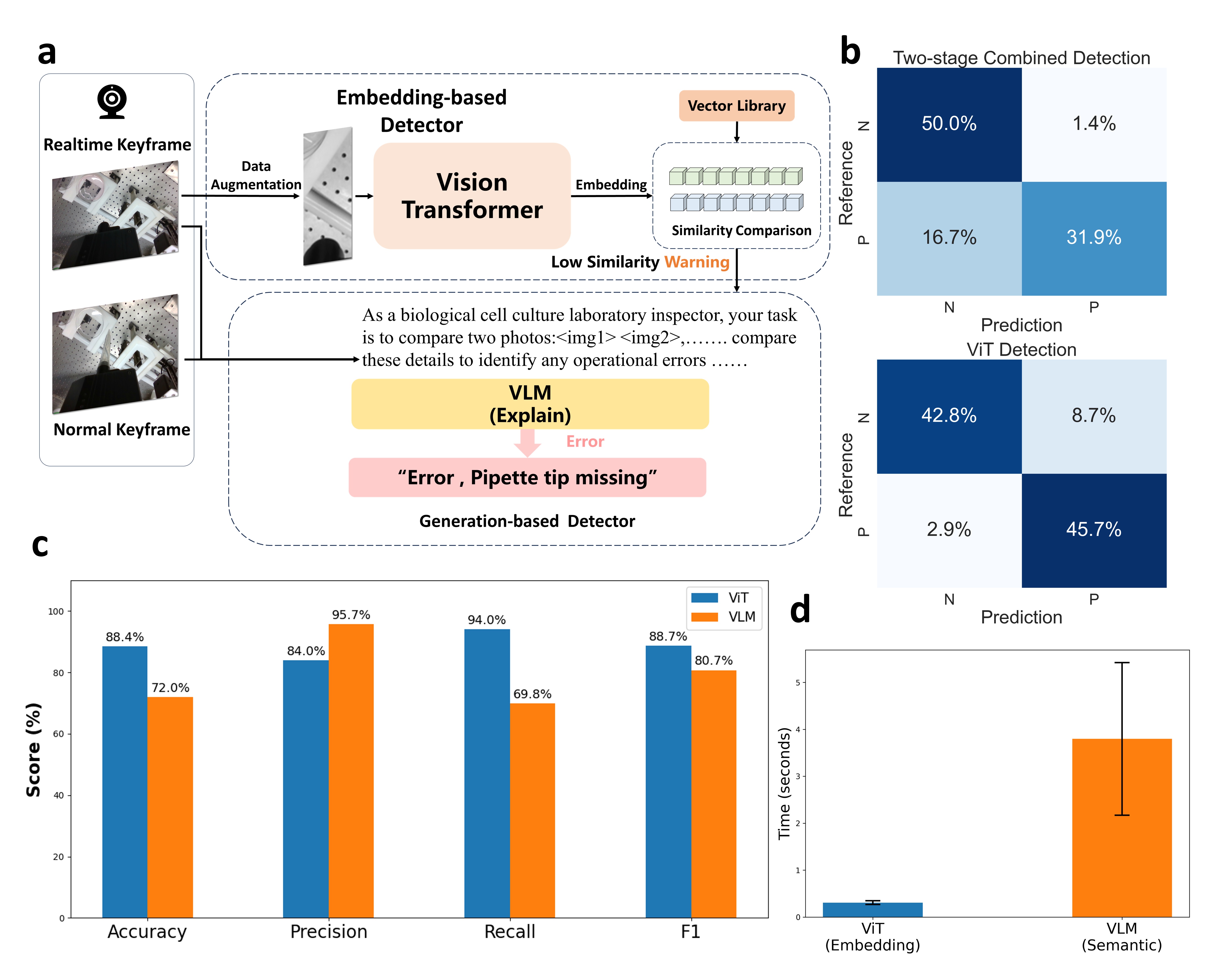}
\caption{\textbf{Inspector Agent Overview and Performance Metrics.}  \textbf{a,} Workflow diagram of the Technician Agent. 
\textbf{b,} Confusion matrix of two-stage combined detection and ViT detection.
\textbf{c,} Performance of ViT and VLM on four evaluation metrics.
\textbf{d,} The time performance of the two detection methods, ViT and VLM.}
\label{fig:inspector_agent}
\end{figure*}

Biological experimentation demands strict precision, where minor procedural errors can compromise outcomes. Conventional automation platforms typically rely on basic object detection without semantic context awareness, limiting their robustness in dynamic laboratory environments \cite{jiang2022review}. To address this, we developed the Inspector Agent—a hierarchical visual monitoring system integrating vision–language models (VLMs) and vision transformers (ViTs) \cite{han2022survey} for multi-stage perception and error detection (Fig.~\ref{fig:inspector_agent}a).

The first stage performs visual segmentation of experimental scenes using few-shot prompting with a VLM. Key objects—such as pipette tips, culture plates, and tubes—are segmented from raw RGB inputs. To enhance spatial resolution and minimize background interference, the bounding boxes generated by the VLM are manually refined. These cropped subregions are converted to grayscale, preserving structural cues like pipette orientation and tube angles while reducing color-based noise.

In the second stage, a ViT-based keyframe detection module encodes 23 visually discriminative actions (selected from 11 control primitives) into a reference embedding library. This module enables sub-second recognition of procedural steps. In benchmark testing, the ViT achieved a mean inference latency of 0.3066 s—91.9\% faster than GPT-4o (3.7960 s)—with lower temporal variability (coefficient of variation: 13.08\% vs. 42.80\%; Fig.~\ref{fig:inspector_agent}d). In real-world experimental settings, the ViT achieved an F1 score of 88.7\% and a recall of 94.0\%, demonstrating high temporal stability and operational fidelity (Fig.~\ref{fig:inspector_agent}c).

A final stage introduces zero-shot semantic validation using the VLM. When the ViT flags anomalies, frames are semantically compared with idealized keyframes using language-guided prompts (e.g., “attach pipette tip”). This semantic differential analysis enables detection of contextual errors beyond geometry alone. In validation, this mechanism achieved 95.7\% precision and 80.7\% F1 score(Fig.~\ref{fig:inspector_agent}c), reducing the false positives rate from 8.7\% to 1.4\%—an 83\% improvement (Fig.~\ref{fig:inspector_agent}b). For example, detecting a detached pipette tip without a visible pipette is correctly flagged as an action violation. Upon confirmation, robotic operations are automatically paused and visual alerts issued.

By integrating geometric and semantic vision processing, the Inspector Agent ensures procedural robustness, accelerates feedback response times, and significantly reduces downstream execution failures.

\subsection{Integrated Biological Experiment Design}
\label{Integrated Biological Experiment Design}

\begin{figure*}[!h]
\centering
\includegraphics[width=0.9\textwidth]{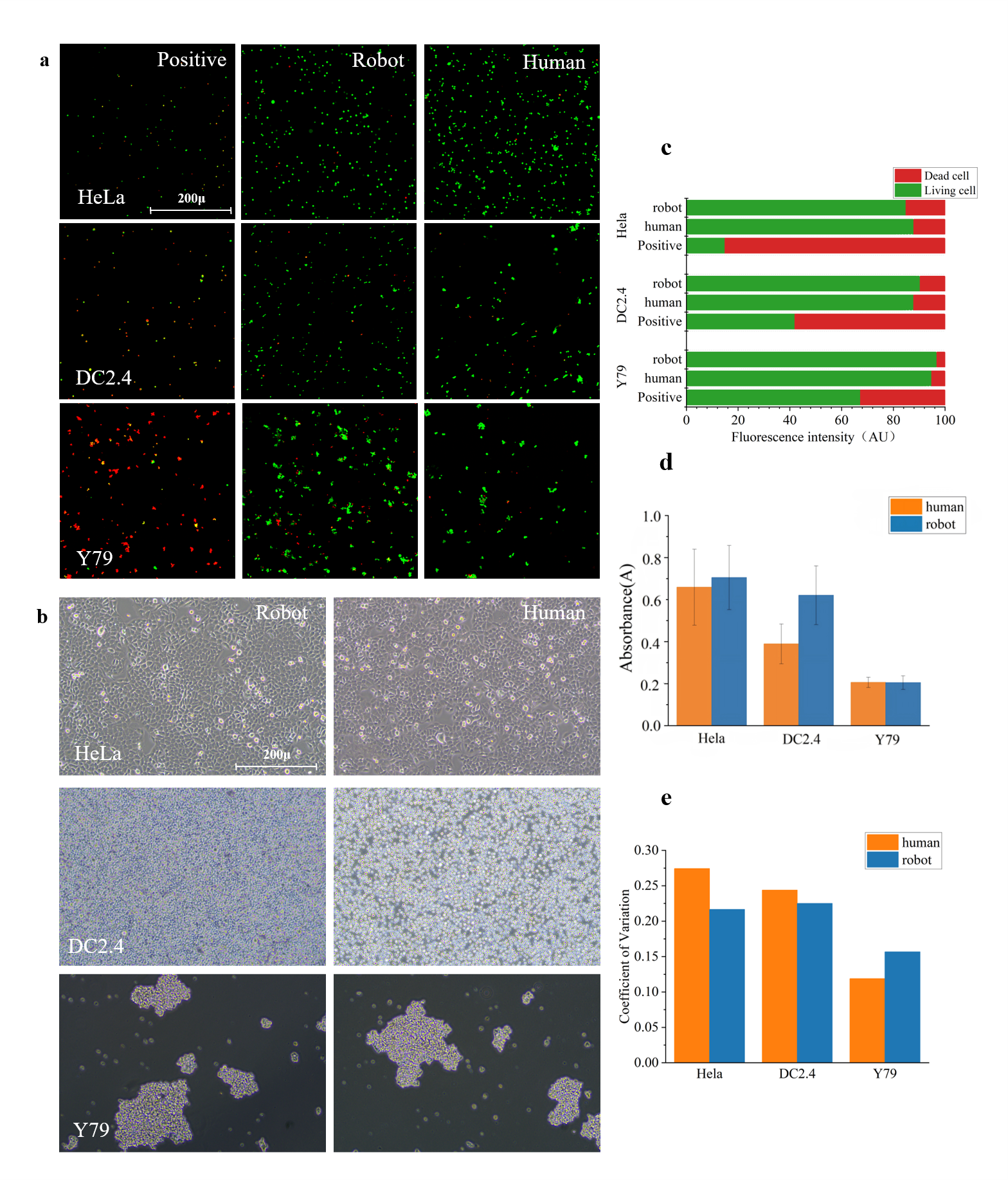}

\caption{\textbf{Comparison of automated vs. manual cell passaging outcomes.}  \textbf{a,} Fluorescence images of live/dead-stained cells (automated vs. manual) at 48 h post-passaging. 
\textbf{b,} Bright-field images of cell morphology post-passaging. 
\textbf{c,} Live/dead cell ratio comparison after passaging.
\textbf{d,} Cell viability comparison between methods.
\textbf{e,} CV of CCK-8 viability across repeats (reproducibility).
}
\label{fig:biological_experiment_result}
\end{figure*}

To evaluate the biological reliability and operational efficiency of BioMARS, we conducted a comparative study between automated and manual cell passaging protocols across three representative cell types: HeLa (adherent), Y79 (suspension), and DC2.4 (semi-adherent/suspension). Experimental evaluation included metabolic viability, survival consistency, morphological preservation, and coefficient of variation (CV) analysis. All workflows adhered to established protocols, with BioMARS dynamically adapting process parameters to each cell line.

Cells were cultured in standard media—HeLa in DMEM with 10\% FBS and 1\% penicillin–streptomycin, Y79 in RPMI-1640 with 20\% FBS, and DC2.4 in RPMI-1640 with 10\% FBS—under 5\% CO₂ at 37 °C. Media changes were performed every 2–3 days. For passaging, adherent cells were detached with 0.25\% trypsin–EDTA. The BioMARS system adjusted enzymatic digestion time and centrifugation based on cell type: 6 minutes for HeLa and 3 minutes for Y79, ensuring optimal yield and viability.

Metabolic viability was assessed 48 hours post-passaging using the CCK-8 assay. Optical density (OD) measurements showed no significant difference between BioMARS and manual protocols across all three cell types (Fig.~\ref{fig:biological_experiment_result}d), indicating that automated processing maintained normal cellular proliferation. CV analysis revealed enhanced reproducibility in the BioMARS group: HeLa and Y79 samples exhibited 12–18\% lower variability compared to manual handling (Fig.~\ref{fig:biological_experiment_result}e).

Live/dead staining confirmed high post-passaging viability, with over 92\% concordance between automated and manual groups (Fig.~\ref{fig:biological_experiment_result}a,b). Green fluorescence indicated dominant live-cell populations, with clear contrast to the red-stained positive control. Morphological evaluation (Fig.~\ref{fig:biological_experiment_result}c) showed no detectable structural abnormalities, further confirming the BioMARS system’s ability to preserve cell integrity.

In addition to biological fidelity, BioMARS markedly improved operational efficiency. Manual passaging required approximately 60 minutes per cell line, whereas the BioMARS system reduced hands-on time to 5–8 minutes—representing a ~90\% reduction. This time savings translates into higher throughput and improved standardization, minimizing human error and procedural variability.

Collectively, these results establish that BioMARS performs comparably or superior to manual protocols in biological outcome metrics while offering significant gains in consistency, reproducibility, and efficiency.

\subsection{Biological Optimizing Capability}
\label{Biological Optimizing Capability}
\begin{figure*}[!h]
\centering
\includegraphics[width=0.9\textwidth]{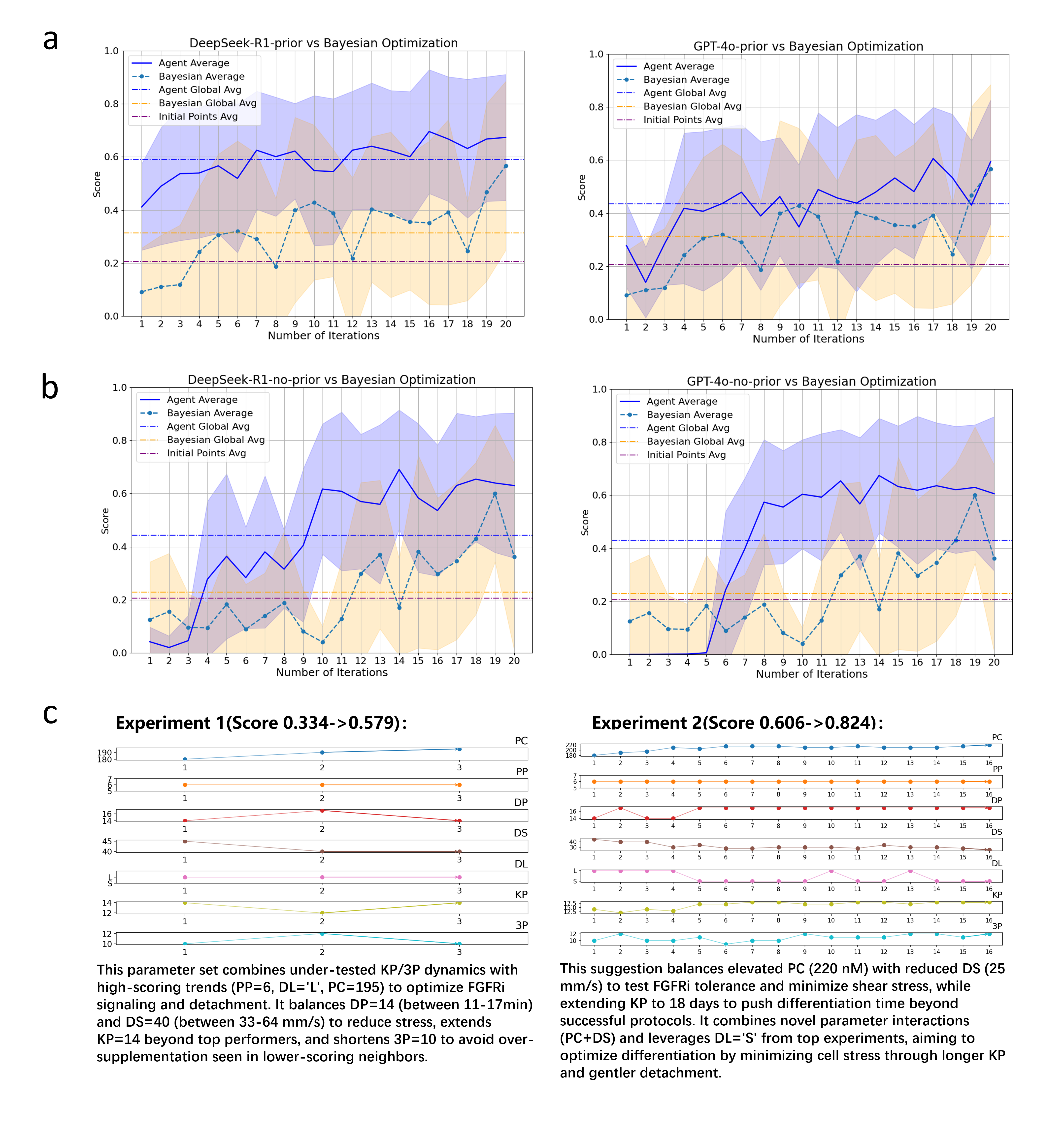}
\caption{\textbf{Results of the iPSC-RPE optimization experiments.}  \textbf{a,} Performance comparison between DeepSeek-R1 and GPT-4o models versus Bayesian Optimization using 10 prior experimental data points. 
\textbf{b,} Comparative analysis of DeepSeek-R1 and GPT-4o against Bayesian Optimization without leveraging prior experimental knowledge. 
\textbf{c,} Parameter recommendations from LLM-based optimizers across selected iteration rounds.}
\label{fig:biological_optimizing}
\end{figure*}

Beyond static protocol generation, the Biologist Agent was evaluated for its capacity to perform biological optimization—an advanced task requiring iterative reasoning, mechanistic understanding, and strategic parameter adjustment. We assessed this capability using a publicly available dataset for optimizing differentiation efficiency of induced pluripotent stem cell-derived retinal pigment epithelial (iPSC-RPE) cells \cite{kanda2022robotic}, which defines a high-dimensional experimental space grounded in biological constraints.

The optimization target was the pigment score, a key phenotypic marker of iPSC-RPE maturation. Seven tunable parameters were considered across preconditioning, detachment, and differentiation stages: FGFRi concentration (PC: 0–505 nM) and exposure duration (PP: 1–6 days); trypsin incubation time (DP: 5–23 min), pipetting strength (DS: 10–100 mm/s), and pipetting length (DL: short/long); KSR withdrawal schedule (KP: 1–19 days); and three-supplement exposure duration (3P: 3–19 days). This setup presents a biologically grounded, combinatorially complex optimization challenge.

To simulate realistic experimental conditions, optimization was constrained to 20 iterations, initialized from 10 randomly selected low-performing conditions (pigment score < 0.6). Parameter selection used KDTree-based nearest-neighbor interpolation \cite{friedman1977algorithm}, with outputs formatted in structured JSON for reproducibility. We compared three strategies: DeepSeek-R1, GPT-4o, and Bayesian optimization under identical initialization settings.

In the prior-informed setting (Fig.~\ref{fig:biological_optimizing}a), DeepSeek-R1 consistently outperformed baselines, reaching a final pigment score of 0.5913—surpassing GPT-4o (0.4344; +15.8\%) and Bayesian optimization (0.3130; +28.5\%). By iteration 7, it achieved 0.6252 and continued steady improvement. GPT-4o plateaued at 0.606, while Bayesian optimization peaked early at 0.5671. DeepSeek-R1’s advantage stems from its ability to encode mechanistic constraints; for instance, in one high-scoring trial (Fig.~\ref{fig:biological_optimizing}c), it selected PC = 220 nM (balancing efficacy and toxicity), DS = 25 mm/s (minimizing shear stress), and KP = 18 days (prolonging Wnt signaling), reflecting domain-consistent reasoning.

GPT-4o occasionally produced viable configurations but lacked consistent convergence, likely due to reliance on pretrained heuristics. Bayesian optimization, devoid of biological priors, frequently proposed implausible combinations (e.g., PC = 405.17 nM; KP = 2 days), resulting in limited progress.

In the no-prior setting (Fig.~\ref{fig:biological_optimizing}b), DeepSeek-R1 again demonstrated robust generalization, reaching performance comparable to the prior-informed case. GPT-4o improved after iteration 8, ultimately reaching a moderate score of 0.6303. Bayesian optimization showed minimal learning, with scores remaining near baseline. DeepSeek-R1 also exhibited superior balance between exploration and exploitation, as evidenced by a lower standard deviation in output scores (0.2366 vs. 0.2447 for GPT-4o and 0.2785 for Bayesian optimization), enabling more stable convergence.

These results validate the potential of knowledge-integrated LLMs to optimize complex biological systems under data-sparse conditions. By combining contextual reasoning with structured decision-making, such agents reduce dependency on manual tuning and offer scalable solutions for experimental design. Future directions include reinforcement learning frameworks to further enhance adaptive feedback integration in regenerative biology workflows.

\section{Discussion}
\label{sec:Discussion}



 
This study introduces BioMARS, an intelligent agent system driven by LLMs and VLMs, capable of autonomously designing, planning and executing biological experiments. By integrating language-driven reasoning with multimodal perception and robotic control, BioMARS addresses the procedural complexity of biological workflows and generates reproducible, high-quality outcomes. These capabilities are enabled by granting LLMs and VLMs access to essential research tools, including scientific literature, programming environments and robotic execution platforms.  The development of such integrated AI systems holds substantial promise for accelerating discovery in the life sciences.

While BioMARS demonstrates robust performance in standard cell culture tasks, several challenges remain. Its operation under atypical or highly customized experimental conditions is limited, with occasional human oversight required for critical steps such as pipetting volumes and centrifugation parameters. Furthermore , although BioMARS integrates multimodal reasoning to interpret and execute experimental protocols, its dependence on existing online procedures limits its capacity for adaptive parameter tuning across diverse laboratory contexts.  Its responsiveness to unexpected experimental deviations is also limited, as real-time judgment remains an open challenge.  Ongoing efforts focus on enhancing the system’s adaptability and fault tolerance using advanced learning algorithms, with preliminary improvements observed.

BioMARS represents a substantial step toward scalable, reproducible automation in biological research.  By addressing key barriers in protocol interpretation and execution, it lays the groundwork for more reliable, flexible and scalable research practices.  Its ability to ensure consistent procedural replication enhances reproducibility and quality control, both of which are critical in applications such as drug discovery and cell model production.  In parallel , automation reduces operational burden, with notable gains in time, material efficiency and labor reduction.  Given that labor constitutes a major cost component in biological production, deployment of BioMARS offers the potential to significantly lower operational expenses. This economic benefit is expected to increase proportionally with production scale.

\section{Method}
\label{sec:Method}

\subsection{Construction of LLM-based and VLM-based agents}\label{Construction of LLM-based and VLM-based agents}

BioMARS is implemented using a modular agent-based architecture, leveraging GPT-4o and the LangChain framework. The Biologist Agent integrates Google Serper and Bing APIs within a dual-engine retrieval system, supporting document acquisition for experimental planning. It applies a Retrieval-Augmented Generation (RAG) strategy, augmented by a Knowledge Checker submodule, to extract biologically relevant information. Retrieved knowledge is combined with task-specific parameters by the Workflow Generator, which constructs executable protocols in natural language. These are refined for logical consistency and constraint adherence by the Workflow Checker.

The Technician Agent converts natural language instructions into executable pseudo-code using a two-stage pipeline. The Code Generator module outputs parameterized function calls representing laboratory actions, which are validated and corrected by the Code Checker module for semantic precision and functional completeness.

The Inspector Agent implements hierarchical visual monitoring. At the perceptual level, InternViT-6B-448px-V1-5 encodes scene features and keyframes via vector similarity for spatial anomaly detection. At the semantic level, a VLM interprets image content using natural language prompts to identify high-level procedural inconsistencies. This layered architecture enables both geometric accuracy and contextual understanding.

\subsection{Web application}
\label{Web application}






The BioMARS human–machine interface is developed with HTML, JavaScript, and CSS within a Flask framework, enabling real-time visualization, alert management, and interactive task control.  Backend logic, written in Python, orchestrates the actions of the Biologist, Technician, and Inspector Agents and manages execution of experimental workflows, including robotic control and device coordination.

Low-latency, bidirectional communication with the robotic platform is achieved via WebSocket protocols, allowing continuous status updates and immediate command execution.  This infrastructure ensures synchronization between the interface, agents, and physical hardware for consistent, responsive experimentation.

\subsection{Modular Robotic System for Automated Protocols}
\label{Modular Robotic System for Automated Protocols}

The hardware platform includes a collaborative dual-arm robot, four modular workstations (for tube and reagent bottle handling, and container manipulation), an automated centrifuge, and a CO₂ incubator. Robotic arms are equipped with grippers and pipette for consumable transfer and liquid handling. container stations include lid actuators and tilting mechanisms (15°), while tube and bottle stations combine grippers with lid actuators. All modules are controlled by STM32 microcontrollers, and the centrifuge communicates over RS-485 for autonomous operation.

The control architecture is built on the Robot Operating System (ROS), with user instructions parsed into ROS Action calls and dispatched via a Client–Service framework. A standardized API of 11 base functions governs motion, pipetting, and manipulation operations, allowing abstraction of hardware-specific controls. This modular structure supports plug-and-play integration of new instrumentation.

\subsection{Architecture of Hierarchical Error Detection}
\label{Architecture of Hierarchical Error Detection}

A two-stage anomaly detection strategy is employed. The first stage applies a ViT-based keyframe detector to identify deviations in experimental operations. Input frames $\mathbf{o}_t$ are segmented, grayscale-normalized, and embedded via $\phi: \mathcal{O}_{\text{processed}} \to \mathcal{F}$ to form feature vectors $f_t$. These are compared with a task-specific reference library $\mathcal{E}$ to compute pairwise similarities:
\begin{equation}
S_{\text{intra}}^{(j,m)} = \left\{\phi_{\text{sim}}(f_p, f_q) \mid \forall p < q\right\},
\end{equation}
with thresholds adaptively selected based on the $\alpha$-quantile of descending scores.

If anomalies are detected, semantic-level validation is triggered. The original frame $\mathbf{o}_t$ is paired with a language interpretation $\mathcal{L}_t = \psi(\mathbf{o}_t)$ and evaluated by the Inspector Agent using task-specific constraints $\mathcal{C}_{\text{task}}$. Anomaly reports are generated as:
\begin{equation}
\mathcal{E}_{\text{VLM}} = f_{\text{error}}(\mathbf{o}_t, \mathcal{C}_{\text{task}}),
\end{equation}
and relayed to users and the Biologist Agent, completing a closed-loop visual reasoning cycle.

\backmatter

\bmhead{Acknowledgements}
We thank Tian Xue for insightful discussions and valuable guidance. This work was supported by the National Key R\&D Program of China (Grant No. 2022YFA1104800) and the Gusu Leading Talent Entrepreneurship and Innovation Program (Grant No. ZL2024349).

\bmhead{Author contributions:}
Y.Qiu. led system development and technical validation, and contributed to system integration and biological experimentation. Z.H. participated in robotic control, control circuitry development, and biological validation. Z.W. contributed to the development and validation of LLM-based workflows. H.L. was responsible for mechanical design and experimental validation. Y.Qiao. contributed to the design and validation of biological experiments. Y.H. was involved in system verification and web interface development. S.S. conducted statistical analysis of biological data. H.P. contributed to the formulation of biological protocols.R.X.X. and M.S participated in the overall experimental planning and supervised the project and experimental framework. Y.Qiu., Z.H., Z.W., H.L., Y.Qiao., Y.H., S.S., R.X.X., and M.S. collectively wrote the manuscript.

\bmhead{Competing Interests}
The authors declare no competing interests.

\bmhead{Data availability}
Examples of the experiments discussed in the text are provided in the Supplementary Information.

\bmhead{Code availability}
Simpler implementation is provided at https://github.com/AlexandreQ27/BioMARS


\bibliography{sn-bibliography}

\begin{appendices}

\section{BioMARS Algorithmic Framework}\label{secA}
\subsection{Core Algorithm Implementation}\label{secA1:Core Algorithm Implementation}
\begin{algorithm}[H]
\caption{Algorithm of BioMARS}\label{alg:emasc}
\begin{algorithmic}
    \State \textbf{Input:}
    \Statex \hspace*{0.5cm} $Q$: User query or problem statement
    \Statex \hspace*{0.5cm} $D_{src}$: Source documents and online resources
    \Statex \hspace*{0.5cm} $M_R$: Robotic manipulation system
    \Statex \hspace*{0.5cm} $E$: Experimental environment
    \Statex \hspace*{0.5cm} $F$: Function library (11 basic actions API)
    \State \textbf{Output:}
    \Statex \hspace*{0.5cm} $R$: Experiment results
    \Statex \hspace*{0.5cm} $L$: Execution log with errors
    \State \textbf{Procedure:}
    %
    \State \textbf{1. Knowledge Acquisition:}
    \Statex \hspace*{0.5cm} $S_{lit} \gets \text{LiteratureSearch}(Q)$ \Comment{Literature search using Google/Bing}
    \Statex \hspace*{0.5cm} $P_{rag} \gets \text{RAG}(S_{lit})$ \Comment{Extract key paragraphs using RAG}
    \Statex \hspace*{0.5cm} $K \gets \text{KnowledgeChecker}(P_{rag})$ \Comment{Reorganize knowledge}
    
    \State \textbf{2. Workflow Generation:}
    \Statex \hspace*{0.5cm} $W_{raw} \gets \text{WorkflowGenerator}(K, Q, E)$ \Comment{Generate initial workflow}
    \Statex \hspace*{0.5cm} $W_{env} \gets \text{WorkflowChecker}(W_{raw}, E)$ \Comment{Adapt to experimental environment}
    
    \State \textbf{3. Code Generation:}
    \Statex \hspace*{0.5cm} $C_{base} \gets \text{CodeGenerator}(W_{env}, F)$ \Comment{Map to 11 basic actions API}
    \Statex \hspace*{0.5cm} $C_{valid} \gets \text{CodeChecker}(C_{base}, F)$ \Comment{Validate logic and interfaces}
    
    \State \textbf{4. Execution with Real-time Anomaly Detection:}
    \For{$a \in \text{Actions}(C_{valid})$}
        \State $M_R \gets \text{StartAction}(a)$ \Comment{Begin executing action $a$}
        
        \While{\text{IsActionRunning}(a)} \Comment{Monitor during execution}
            \State $e_{vit} \gets \text{ViTKeyPointDetection}(\text{CurrentState}(a))$ \Comment{Keyframe visual detection}
            \If{$e_{vit} = \text{warning}$}
                \State $e_{vlm} \gets \text{VLMKeyPointDetection}(\text{CurrentState}(a))$ \Comment{Trigger VLM for semantic analysis}
                \If{$e_{vlm} = \text{error}$}
                    \State $L \gets L \cup e_{vlm}$ \Comment{Log semantic error}
                    \State \text{Alert}($e_{vlm}$) \Comment{Trigger user alert}
                    \State \text{AbortAction}(a) \Comment{Immediately stop execution}
                    \State \textbf{goto} \text{Step 5} \Comment{Replanning required}
                \EndIf
            \EndIf
        \EndWhile
        
        \State $M_R \gets \text{CompleteAction}(a)$ \Comment{Action completed successfully}
    \EndFor
    
    \State \textbf{5. Replanning:}
    \Statex \hspace*{0.5cm} $W_{new} \gets \text{BiologistPlanner}(L, W_{env})$ \Comment{Human-in-the-loop planning}
    \Statex \hspace*{0.5cm} \text{Goto} \text{Step 3} 
    
    \State \textbf{6. Return:}
    \Statex \hspace*{0.5cm} \Return $R, L$
\end{algorithmic}
\end{algorithm}

\section{Biologist Agent}\label{secB}
\subsection{10-Query Biological Task Set}\label{secB1:10-Query Biological Task Set}
The following 10 standardized questions apply to \textbf{all 7 cell lines} (HeLa, HUVEC, HepG2, DC2.4, Y79, K562, CHO). Each question uses the same format, with the only variation being the **cell type name**:

\begin{enumerate}
    \item \textbf{How to resuscitate [Cell Type] cells in detail?}
    \item \textbf{How to perform passaging of [Cell Type] cells in detail?}
    \item \textbf{How to change the medium for [Cell Type] cells in detail?}
    \item \textbf{How to freeze and store [Cell Type] cells in detail?}
    \item \textbf{What is the recommended seeding density for the [Cell Type] cell line?}
    \item \textbf{How to detect the metabolic activity of the [Cell Type] cell line in detail?}
    \item \textbf{How to evaluate the apoptotic level of the [Cell Type] cell line in detail?}
    \item \textbf{What is the cryopreservation solution formula for the [Cell Type] cell line?}
    \item \textbf{How to detect the proliferation of the [Cell Type] cell line in detail?}
    \item \textbf{How to culture 3D cell spheres using the [Cell Type] cell line in detail?}
\end{enumerate}

\subsection{Protocol Generation Evaluation Metrics}\label{secB2:Protocol Generation Evaluation Metrics}
\subsection*{Score 5}
\textbf{User Question}: \emph{"How to change the medium for HepG2 cells in detail?"} \\
\textbf{Model}: deepseek-r1+WC+KC \\
\textbf{Standard}: Very detailed and biologically accurate cell culture procedure description \\
\textbf{Steps}:
\begin{enumerate}
    \item Retrieve ContainerA from the incubator and place it on one of the two pipetting platforms.
    \item Aspirate 10ml of old medium using a sterile pipette.
    \item Add 10ml pre-warmed (37°C) PBS, gently shake horizontally for 10 seconds, then aspirate completely.
    \item Add 10ml fresh pre-warmed (37°C) culture medium (e.g., DMEM with 10\% FBS), shake horizontally for 10 seconds to distribute.
    \item Return ContainerA to the incubator (37°C, 5\% CO₂).
\end{enumerate}

\textbf{Reason}:
\begin{itemize}
    \item Includes precise volumes (10ml PBS/10ml medium).
    \item Specifies reagent pre-warming (37°C) and explicit culture conditions (37°C/5\% CO₂).
    \item Detailed operational instructions (e.g., horizontal shaking).
\end{itemize}

\subsection*{Score 4}
\textbf{User Question}: \emph{"How to change the medium for HepG2 cells in detail?"} \\
\textbf{Model}: deepseek-r1+WC \\
\textbf{Standard}: Detailed and biologically accurate but lacks reagent quantities or parameter anomalies \\
\textbf{Steps}:
\begin{enumerate}
    \item Pre-warm 10ml PBS and 10ml fresh medium to 37°C.
    \item Retrieve ContainerA, aspirate 10ml old medium using a serological pipette.
    \item Add 10ml PBS, gently swirl for 10 seconds, aspirate. Repeat PBS wash.
    \item Add 10ml fresh medium, swirl in a figure-eight motion for 15 seconds.
    \item Return ContainerA to the incubator.
\end{enumerate}

\textbf{Reason}:
\begin{itemize}
    \item Steps are operationally correct but omit critical medium composition details (e.g., "DMEM + 10\% FBS").
    \item Volumes are specified, but the lack of reagent specifics reduces biological clarity.
\end{itemize}

\subsection*{Score 3}
\textbf{User Question}: \emph{"How to freeze and store HUVEC cells in detail?"} \\
\textbf{Model}: gpt-3.5-turbo+WC \\
\textbf{Standard}: The correct biological cell culture is carried out, but there are logical errors in the steps \\
\textbf{Steps}:
\begin{enumerate}
    \item Take out ContainerA from the incubator.
\item Discard old culture dish.
\item Obtain new culture dishes ContainerB and ContainerC.
\item Place ContainerA on a pipetting platform.
\item Add 5ml of liquid to ContainerA.
\item Add 2ml of enzyme solution to ContainerA.
\item Shake ContainerA to ensure even distribution of the enzyme solution.
\item Centrifuge ContainerA in TubeA at 1000 x g for 5 minutes.
\item Remove the liquid after centrifugation.
\item Add 5ml of culture medium to ContainerA.
\item Distribute the cell suspension into ContainerB and ContainerC in perfect aliquots.
\item Shake ContainerB and ContainerC to ensure uniform dispersion of cells.
\item Place ContainerB on a pipetting platform.
\item Add 5ml of culture medium to ContainerB.
\item Store ContainerB in the incubator for cryopreservation.
\item Place ContainerC on a pipetting platform.
\item Add 5ml of culture medium to ContainerC.
\item Store ContainerC in the incubator for cryopreservation.
\end{enumerate}

\textbf{Reason}:
\begin{itemize}
    \item Specific details are lacking, and there are issues with parameters such as centrifugal force.
\end{itemize}

\subsection*{Score 2}
\textbf{User Question}: \emph{"How to change the medium for HepG2 cells in detail?"} \\
\textbf{Model}: deepseek-r1\\
\textbf{Standard}: Extremely vague or infeasible description \\
\textbf{Steps}:
\begin{enumerate}
    \item Remove ContainerA, aspirate old medium.
    \item Add 5ml PBS, swirl, aspirate.
    \item Add 3ml trypsin-EDTA, incubate 5 minutes.
    \item Neutralize with 6ml medium, transfer to TubeA.
    \item Centrifuge TubeA at 300xg for 5 minutes, resuspend in 10ml fresh medium.
    \item Transfer 5ml to ContainerB and ContainerC, incubate.
\end{enumerate}

\textbf{Reason}:
\begin{itemize}
    \item Includes irrelevant steps for HepG2 medium change (e.g., trypsin digestion, centrifugation, subculturing).
    \item Logic is flawed (e.g., splitting into multiple containers). HepG2 medium change does not require cell passaging.
\end{itemize}

\subsection*{Score 1}
\textbf{User Question}: \emph{" How to resuscitate HeLa cells in detail?"} \\
\textbf{Model}: gpt-3.5-turbo \\
\textbf{Standard}: Incorrect answers or failure to follow instructions \\
\textbf{Steps}:
\begin{enumerate}
    \item Take out ContainerA from the incubator.
\item Obtain two new 10cm cell culture dishes named ContainerB and ContainerC from the petri dish storage rack.
\item Add 10ml of culture medium to ContainerB and ContainerC.
\item Discard the old culture dish ContainerA.
\item Add 5ml of cell suspension to ContainerB and ContainerC in perfect aliquots.
\item Shake the culture dishes to ensure even distribution of cells.
\item Place the culture dishes on the pipetting platform.
\item Add 2ml of enzyme and buffer solution to ContainerB and ContainerC.
\item Shake the culture dishes again to ensure uniform dispersion of cells in the cell suspension.
\item Centrifuge the culture dishes at a centrifugal force of 3000g for 5 minutes using TubeA.
\item Remove the liquid used to wash the cells after washing.
\item Resuspend the cells in ContainerB and ContainerC by adding 10ml of fresh culture medium.
\item Place the culture dishes back in the incubator for further growth.
\end{enumerate}

\textbf{Reason}:
\begin{itemize}
    \item It is completely unclear what it is doing.
\end{itemize}

\section{Technician Agent}\label{secC}
\subsection{Functional Interface Specifications}\label{secC1:Functional Interface Specifications}
\begin{longtable}{|m{4.5cm}|m{4cm}|m{4cm}|}
    \caption{Function List for Cell Culture Operations} \\
    \hline
    \textbf{Function} & \textbf{Description} & \textbf{Parameters} \\
    \hline
    \endfirsthead
    \hline
    \textbf{Function} & \textbf{Description} & \textbf{Parameters} \\
    \hline
    \endhead

    \texttt{take\_out\_cells(list$<string>$ containers)} & 
    Remove the culture dishes containing cells that need to be used from the incubator. & 
    containers: A list of identifiers for the culture containers containing the cells to be removed. \\
    \hline
    \texttt{put\_back\_incubator(list$<string>$ containers, int detachment\_time=0)} & 
    Place the culture dishes back into the incubator. If the cells have been treated with an enzyme for detachment, this function ensures proper handling and incubation conditions for the detachment process. & 
    containers: A list of identifiers for the culture dishes to be returned to the incubator; detachment\_time: The time in minutes for which the cells should remain in the incubator for detachment (default is 0). \\
    \hline
    \texttt{remove\_liquid(float volume, string container)} & 
    Use a pipette tip to aspirate and remove liquid from the designated culture container. & 
    volume: Volume of liquid to aspirate (in ml); container: The designated culture container number. \\
    \hline
    \texttt{add\_liquid(string liquid\_type="PBS", float volume, string container)} & 
    Aspirate a specified solution and add it to the designated container. & 
    liquid\_type: The type of the specified solution (default: "PBS"); volume: Volume of solution to be added (in ml); container: The designated container identifier to which the liquid is to be added. \\
    \hline
    \texttt{detach\_cells\_with\_pipette (string container)} & 
    After enzymatic treatment and neutralization, use a pipette to gently detach and resuspend the cells in the existing liquid within the culture container. & 
    container: The identifier for the culture container containing the enzyme-treated cells. \\
    \hline
    \texttt{shake(string container)} & 
    Gently shake the designated culture container to ensure mixing or resuspension. Typically performed after adding enzymes, buffers, or during washing. & 
    container: The identifier of the culture container to be shaken. \\
    \hline
    \texttt{centrifuge(int speed, int time, string container)} & 
    Centrifuge a tube at a specified speed for a set duration. & 
    speed: Rotational speed of the centrifuge (in g); time: Duration of centrifugation (in minutes); container: Identifier of the container holding the centrifuge tube. \\
    \hline
    \texttt{resuspension(string container)} & 
    Resuspend cells in a centrifuge tube by aspirating and dispensing the liquid. & 
    container: Identifier of the centrifuge tube. \\
    \hline
    \texttt{remove\_supernatant(string container)} & 
    Pour out the supernatant from the centrifuge tube without disturbing the cell pellet. & 
    container: Identifier for the culture container. \\
    \hline
    \texttt{get\_container(string container)} & 
    Pick up a specified culture container from its storage rack. & 
    container: Unique identifier for the container. \\
    \hline
    \texttt{discard\_container(string container)} & 
    Discard or relocate a used culture container. & 
    container: Unique identifier for the container. \\
    \hline
\end{longtable}

\section{Inspector Agent}\label{secD}
\subsection{Error Scenario}\label{secD1:Error Scenario}
\begin{longtable}{|m{1.5cm}|m{11.5cm}|}
    \hline
    \textbf{Error ID} & \textbf{Description} \\
    \hline
    \endfirsthead
    \hline
    \textbf{Error ID} & \textbf{Description} \\
    \hline
    \endhead
    1 & During liquid removal: A pipette tip is inserted, but no cell culture dish is placed on the platform \\
    \hline
    2 & During liquid removal: No pipette tip is installed \\
    \hline
    3 & During liquid removal: The cell culture dish lid is not properly closed \\
    \hline
    4 & During liquid removal: The platform is not lifted during liquid aspiration, preventing the pipette from reaching the dish \\
    \hline
    5 & During liquid addition: No pipette tip is installed \\
    \hline
    6 & During liquid addition: A pipette tip is installed but reagent bottle 1 is not connected \\
    \hline
    7 & During liquid addition: The lid of reagent bottle 1 is not opened \\
    \hline
    8 & During liquid addition: The dish lid is not opened before aspirating from container A \\
    \hline
    9 & During liquid addition: The pipette tip detaches during aspiration from container A \\
    \hline
    10 & During liquid addition: Test tube A is not placed during aspiration \\
    \hline
    11 & During liquid addition: The pipette tip detaches during aspiration from test tube A \\
    \hline
    12 & During liquid addition: The platform is not lifted during liquid ejection into the dish, causing overflow \\
    \hline
    13 & During liquid addition: The pipette tip detaches during liquid ejection into the dish \\
    \hline
    14 & During centrifugation: The centrifuge rotor is not in place before inserting tubes, causing the tubes to fail to be placed \\
    \hline
    15 & During centrifugation: No centrifuge tubes are prepared \\
    \hline
    16 & During centrifugation: The rotor is not in place after centrifugation, preventing tube removal \\
    \hline
    17 & During resuspension: No pipette tip is installed \\
    \hline
    18 & During resuspension: The pipette tip falls off during resuspension \\
    \hline
    19 & During cell pipetting: No pipette tip is installed before pipetting \\
    \hline
    20 & During cell pipetting: The pipette tip detaches during pipetting \\
    \hline
\end{longtable}
\label{tab:errors}

\clearpage

\section{Robotic Execution Platform}\label{secE}
\subsection{Hardware}\label{secE1:Hardware}
\begin{figure*}[htbp]
\centering
\includegraphics[width=0.9\textwidth]{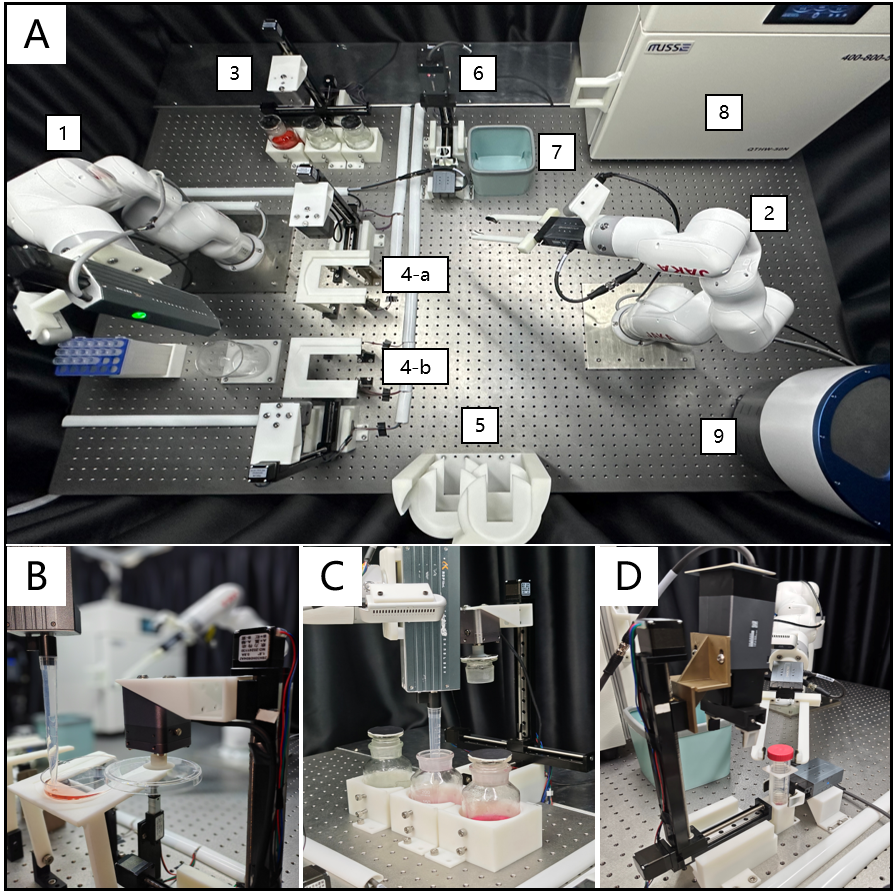}

\caption{\textbf{Hardware configuration of the BioMARS platform.} 
\textbf{A}, Robotic workspace layout designed to ensure all custom devices and standard laboratory equipment are within the reach of both arms. 
(1) Pipetting arm; 
(2) Gripper arm; 
(3) Automated reagent bottle station with three positions; 
(4-a, 4-b) Petri dish handling stations; 
(5) Petri dish storage rack; 
(6) Automated centrifuge tube capper/decapper station; 
(7) Waste container; 
(8) CO\textsubscript{2} incubator; 
(9) Centrifuge. 
\textbf{B}, Petri dish handling station. 
\textbf{C}, Reagent bottle station with automated capping and uncapping. 
\textbf{D}, Centrifuge tube handling and capper/decapper station.}

\label{fig:hardware}
\end{figure*}

\section{User Interaction Interface}\label{secF}
\subsection{Web-Based Control Panel Design}\label{secF:Web-Based Control Panel Design}

\begin{figure*}[!h]
\centering
\includegraphics[width=0.9\textwidth]{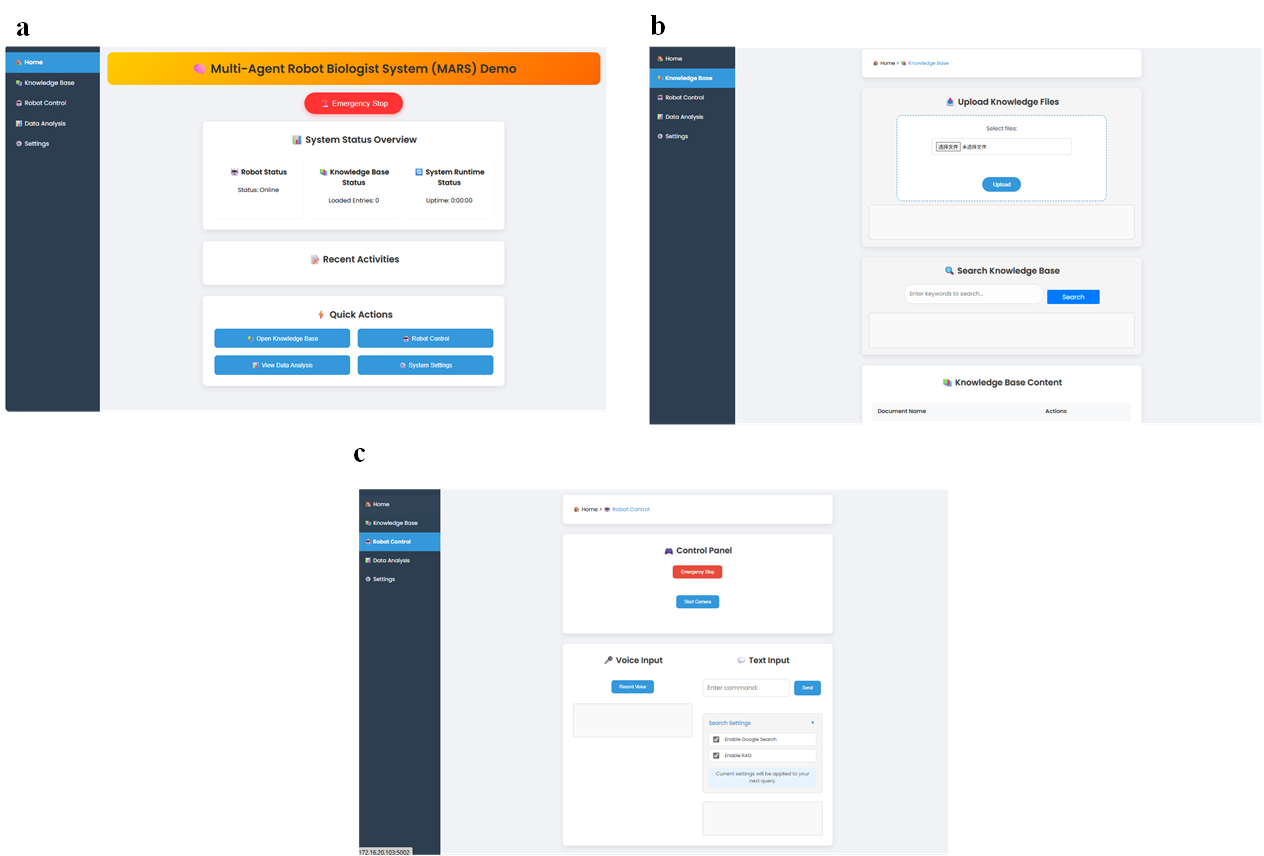}

\caption{\textbf{BioMARS User Interface.}  
\textbf{a,} Home Page Layout. 
\textbf{b,} Knowledge Base Interface. 
\textbf{c,} Robot Control Interface.}
\label{fig:web}
\end{figure*}
The system employs a web-based control panel with a two-column layout.The left navigation menu (fixed on the left side) provides quick access to different functional pages, while the right content area dynamically displays page-specific modules. This design ensures intuitive operation and efficient task execution. (See Fig.\ref{fig:web}a)

The Home Page (index.html) serves as the central dashboard. It features a system title at the top, an emergency stop button for critical operations, and a status overview section (currently under development) that will display real-time parameters like CPU temperature and memory usage. A recent activities timeline (also under development) tracks user interactions, and a quick access area offers shortcuts to robot control, knowledge base, and settings.

The Robot Control Page (robot\_control.html) focuses on operational management.  
A control panel includes emergency stop and camera control buttons for managing robotic arm actions and multi-angle video feeds. The video streaming area allows users to monitor cell culture processes from multiple perspectives. Input controls support voice/text commands for step-by-step instructions, while the action steps panel and system log provide real-time feedback on operations. (See Fig.\ref{fig:web}b)

The Knowledge Base Management Page (knowledge\_base.html) enables document handling. Users can upload files (PDF/DOCX/TXT) for automatic vectorization processing.  A search function supports semantic queries with downloadable results, and the content display area lists all documents in the knowledge base with options for sorting, previewing, and editing.  (See Fig.\ref{fig:web}c)




\end{appendices}

\end{CJK}
\end{document}